\documentclass{article}

\usepackage{arxiv}

\usepackage[utf8]{inputenc} 
\usepackage[T1]{fontenc}    
\usepackage{hyperref}       
\usepackage{url}            
\usepackage{booktabs}       
\usepackage{amsfonts}       
\usepackage{nicefrac}       
\usepackage{microtype}      
\usepackage{lipsum}		
\usepackage{graphicx}

\title{DenseAttentionSeg: Segment Hands from Interacted Objects Using Depth Input}

\date{} 					

\author{
  Zihao Bo \\
  Tsinghua University\\
  \texttt{bzh17@mails.tsinghua.edu.cn} \\
   \And
  Hao Zhang \\
  Tsinghua University\\
  \texttt{zhanghao16@mails.tsinghua.edu.cn} \\
   \AND
  Junhai Yong \\
  Tsinghua University \\
  \texttt{yongjh@tsinghua.edu.cn} \\
   \And
  Feng Xu \\
  Tsinghua University \\
  \texttt{feng-xu@tsinghua.edu.cn} \\
}

\begin{document}
\maketitle

\begin{abstract}
  We propose a real-time DNN-based technique to segment hand and object of interacting motions from depth inputs. Our model is called DenseAttentionSeg, which contains a dense attention mechanism to fuse information in different scales and improves the results quality with skip-connections. Besides, we introduce a contour loss in model training, which helps to generate accurate hand and object boundaries. Finally, we propose and release our InterSegHands dataset \footnote{\url{https://interseghands.meteorshub.com}}, a fine-scale hand segmentation dataset containing about 52k depth maps of hand-object interactions. Our experiments evaluate the effectiveness of our techniques and datasets, and indicate that our method outperforms the current state-of-the-art deep segmentation methods on interaction segmentation.

\end{abstract}


\section{Introduction}
Hand is an important part of our body that makes human be able to construct, manipulate tools and express ideas. Many researches, such as pose recognition, 3D reconstruction, human-computer interaction, AR/VR, etc., have paid a lot of attentions on human hands. Among almost all those works, it is important to detect and segment hands from their interacting objects and backgrounds, and hand-object segmentation can also be applied to many other high-level tasks. 

\begin{figure}[b]
  \begin{center}
      \includegraphics[width=0.55\linewidth]{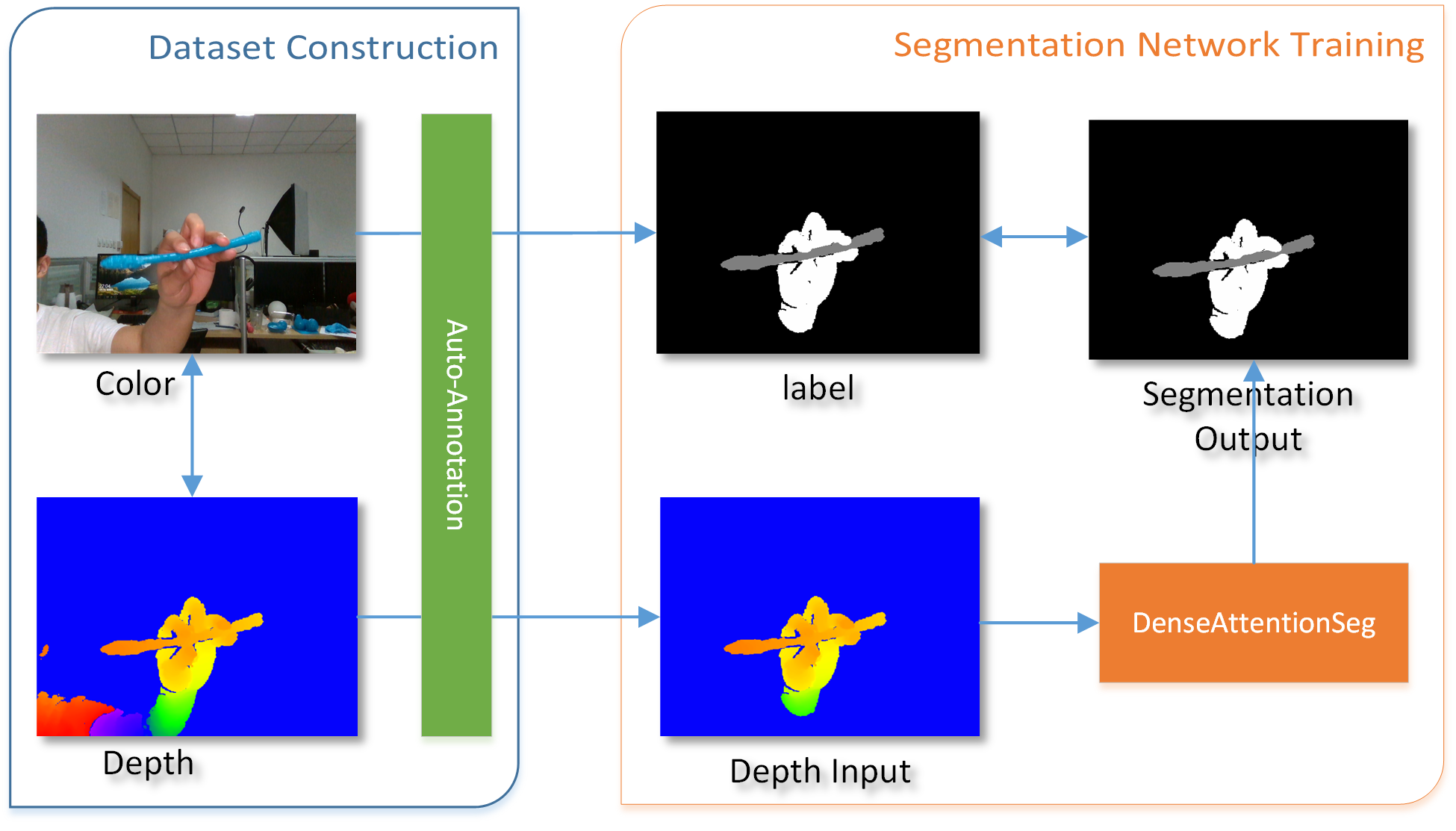}
      \end{center}
  \caption{The workflow of our work. The left blue container indicates our auto-annotation of dataset construction from aligned color and depth maps. The right orange container shows our DenseAttentionSeg model to segment hand from its interacting objects.}
  \label{fig:workflow}
\end{figure}

Many existing works consider RGB images as input. However, RGB images are easily affected by the hand skin, the color of interacting objects and the illumination condition, which add inherent difficulties for hand-related applications. In this paper, we consider using depth images which are more robust to these factors, and become easier to be obtained as more and more consumer depth sensors are available even on mobile devices. And we propose a depth-based hand-object segmentation dataset to promote researches on this topic.

Existing hand segmentation datasets on depth either lack object interaction \cite{tompson2014real} or are almost in large scenes with low-resolution hands and indistinct boundaries \cite{kang2017hand}. A large dataset with high-resolution hand-object interaction and delicate pixel-level labels is eagerly required to meet the demand for network training in the deep learning era. So we propose InterSegHands to match the requirement and accompanying an auto-annotation method which is used to construct the current dataset and can be used to easily enlarge the dataset in the future. We will release the dataset and the auto-annotation method to promote researches in this area.

Nowadays, deep learning methods have been adopted in general semantic segmentation successfully \cite{long2015fully,ronneberger2015u,chen2018deeplab}. In this work, we evaluate some famous deep segmentation models. Furthermore, we propose DenseAttentionSeg, which contains a dense attention structure to fuse the information of different scales and guide the skip-connections. Attention mechanism has been proven effective in CV and NLP, which often computes the attention weights of the encoder in seq-to-seq learning or different regions of the input feature map. Recently, similar ideas are also tried in general semantic segmentation \cite{islam2017gated,li2018pyramid}, but they are limited to high-level guidance. Nevertheless, we build an attention mechanism on different feature scales which not only convey coarser semantic but also finer shape and edge information, which we call it dense attention. In addition, we propose a new contour loss which emphasizes the object contours and promotes the final results a lot, which is particularly designed for this fine-scale hand-object segmentation job and will not decrease the inference speed.

The whole pipeline is shown in Figure \ref{fig:workflow}. Our contributions can be summarized as follows: \textbf{First}, we introduce a large fine-scale dataset for hand segmentation with object interaction using depth maps, together with an auto-annotation method; \textbf{Second}, we propose an encoder-decoder architecture with a dense attention mechanism for this hand segmentation task. As far as we know, this is the first depth-based reconstruction-level segmentation work for hand-object interaction. It outperforms the state-of-the-art semantic segmentation models on our dataset; \textbf{Third}, we present a contour refinement strategy that significantly improves the accuracy of our segmentation result.

\section{Related Work}

As a basic task in hand research, hand segmentation contributes to many hand applications, such as pose estimation \cite{doliotis2012hand,sharp2015accurate}, gesture recognition \cite{Baraldi_2014_CVPR_Workshops,kang2015real} and so on.

\textbf{Color and depth:} As for RGB-based methods, Bambach \emph{et al.} segmented hands egocentrically in many activities via the combination of traditional and neural methods \cite{bambach2015lending}. Urooj \emph{et al.} explored egocentric hand segmentation in social activities \cite{urooj2018analysis}. There are also some hand works (including hand tracking and gesture recognition) based on depth camera \cite{moon2018v2v,kang2015real}. Some methods used random decision forest(RDF) \cite{tompson2014real,sharp2015accurate}. Kang \emph{et al.} adopted a simple segmentation method by using a black wristband in their hand tracking work \cite{kang2015efficient}. They also studied the problem of hand segmentation recently \cite{kang2017hand}.

\textbf{Hand and object interaction:} Though the most common situation of hands in our daily life is to interact with all kinds of objects, it has not been fully researched. Kang \emph{et al.} proposed a two-stage RDF to segment hand and object on depth \cite{kang2017hand}. Their solution runs fast while with a fair accuracy. And most of their dataset samples lack closeup information of hand and object details. Another relevant work is Urooj \emph{et al.}'s \cite{urooj2018analysis}, in which they used a high-performance network to segment hand. But they adopted RGB images as input.

\textbf{Datasets:} Bambach \emph{et al.} introduced EgoHands for hand tasks including segmentation based on RGB images \cite{bambach2015lending}, including 4.8k frames in total. Zimmermann and Brox recently proposed a larger dataset for color images, which has about 44k samples \cite{zimmermann2017learning}. But it is synthetic and difficult to apply in the real world. As for the depth based dataset, Tompson \emph{et al.} released NYU Hand Pose Dataset, which contains about 6k depth frames but lacks object interaction \cite{tompson2014real}. Kang \emph{et al.} collected a total number of 27k depth images for segmentation, but most of their dataset samples do not show the fine structure of hands and lost the object information to some degree \cite{kang2017hand}.

\textbf{Network based semantic segmentation:} Long \emph{et al.} first proposed fully convolutional neural networks \cite{long2015fully}. Ronneberger \emph{et al.} designed an encoder-decoder architecture that was widely used in the following years \cite{ronneberger2015u}. Other methods include PSPNet \cite{zhao2017pyramid}, RefineNet \cite{lin2017refinenet}, Large Kernel Matters \cite{peng2017large}, Deeplab \cite{chen2018deeplab}, and so on. Though existing network-based models work well, they were built for general segmentation tasks and may not all act the same for depth-based hand-object interaction segmentation task where the classes are limited and the hand and object are closely related.

\section{InterSegHands Dataset}

Our database contains depth images of various hand-object interaction motions and the corresponding hand masks of the depth images. To obtain the ground truth segmentation masks, we use the aligned color images. 

We collect depth images of hands using Intel RealSense SR300. We keep the max depth image size of $640\times480$ for high-resolution requirements. We record images under a fixed illumination intensity and color temperature for a better segmentation of the color images. 

\begin{figure}[ht]
    \begin{center}
    \includegraphics[width=0.65\linewidth]{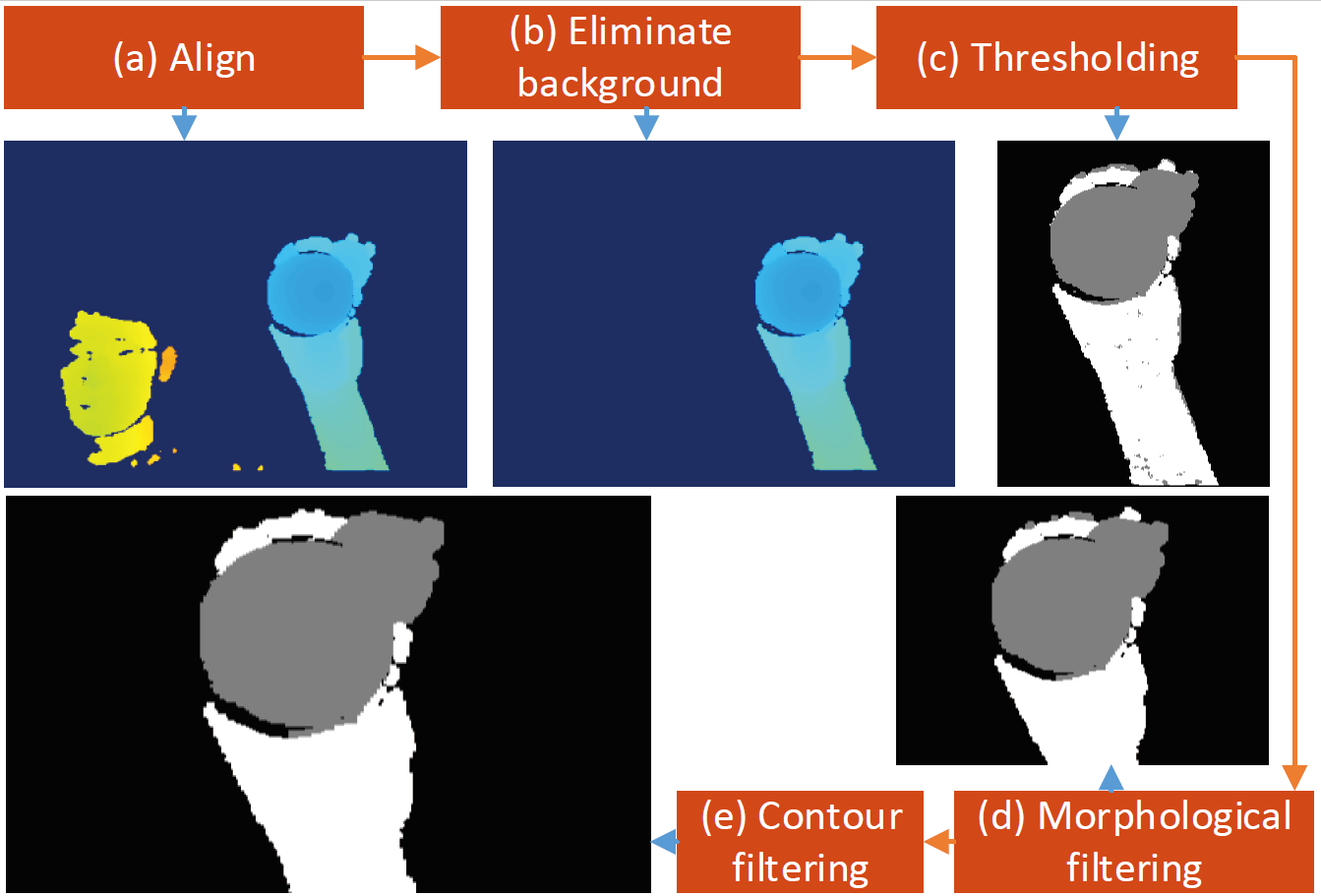}
    \end{center}
       \caption{The procedure of auto-annotation for building the ground truth. Contour filtering can fix the band on top of fingers.}
       \label{fig:auto-annotation} 
\end{figure}

Figure \ref{fig:auto-annotation} shows the procedure of our auto-annotation for building the ground truth of depth image segmentation. \\
\textbf{Background Elimination.} After aligning the color and depth images, we crop the hand area in a fixed depth range of 16cm and discard the background. Here we assume that the hand and object are the nearest parts to the camera. \\
\textbf{Segmentation.} Color space thresholding is able to segment skin from different colored objects. Here we convert color images into HSV space and analyze the linear thresholding restrictions. To reduce noises, we adopt morphological filtering on binary masks of hand and objects respectively. Apart from that, as the alignment cannot be perfect, we adopt a contour filtering step:

\begin{equation}
  \label{contour-filtering}
    M_o=\bigcup\{s\in \mathcal{T}(M_i)|N(s)>\theta, \frac{(\partial N(s))^2}{N(s)}<\phi\}
\end{equation}
Where $M_i$ and $M_o$ are the input and output masks. $s$ is a segment in the input mask, and $\mathcal{T}$ computes all the segments. The area and perimeter of a segment are represented by $N(s)$ and $\partial N(s)$. In our implementation, $N(s)$ is measured by the number of pixels in $s$ while $\partial N(s)$ is the number of boundary pixels. $\theta$ and $\phi$ are the two thresholds. 

Finally, we check the segmentation results for each frame and discard frames with visible errors, which is greatly faster than pixel-wise manual annotation. The experiment shows over 88\% of our auto-annotation results are considered to be acceptable.

\begin{table*}[h]
  \begin{center}
  \begin{tabular}{cccccccc}
  \toprule
  Dataset & Type & Annotation & \# Frames & Hand Resolution & Object Interaction \\
  \midrule
  EYTH\cite{urooj2018analysis} & color & manual & 1,290 & Uncertain & Partial \\
  NYU\cite{tompson2014real} & depth & automatic & 6,736 & Middle & No \\
  HOI\cite{kang2017hand} & depth & - & 27,525 & Low & Yes \\
  \textbf{InterSegHands(Ours)} & depth & automatic & 51,912 & High & Yes \\
  \bottomrule
  \end{tabular}
  \end{center}
  \caption{Statistics of several hand segmentation datasets.}
  \label{table:dataset-summary}
\end{table*}

\begin{figure}[ht]
    \begin{center}
    \includegraphics[width=0.55\linewidth]{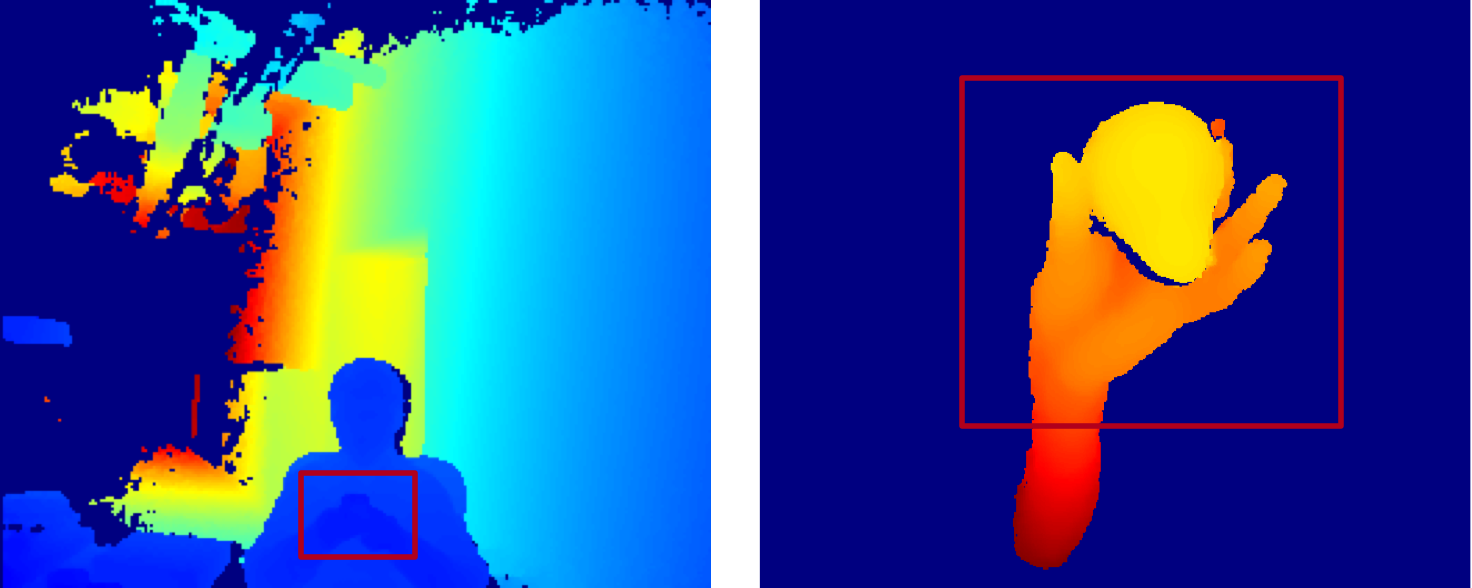}
    \end{center}
       \caption{Left: input sample from a recent relevant work. Right: input sample from ours. The interested area is showed in red box.}
       \label{fig:dataset} 
\end{figure}

We collect a total of 51,912 pairs of depth maps and corresponding ground truth masks with 28 different interacting objects. The whole dataset is split into 46,720 pairs for training and 5,192 for validation. We collect the data in a third person perspective and Table \ref{table:dataset-summary} shows the statistics of our dataset as well as some other hand segmentation datasets. Figure \ref{fig:dataset} shows our high resolution of hand area compared to the recent work \cite{kang2017hand}.

\section{DenseAttentionSeg}

\begin{figure*}[h]
  \begin{center}
  \includegraphics[width=0.95\linewidth]{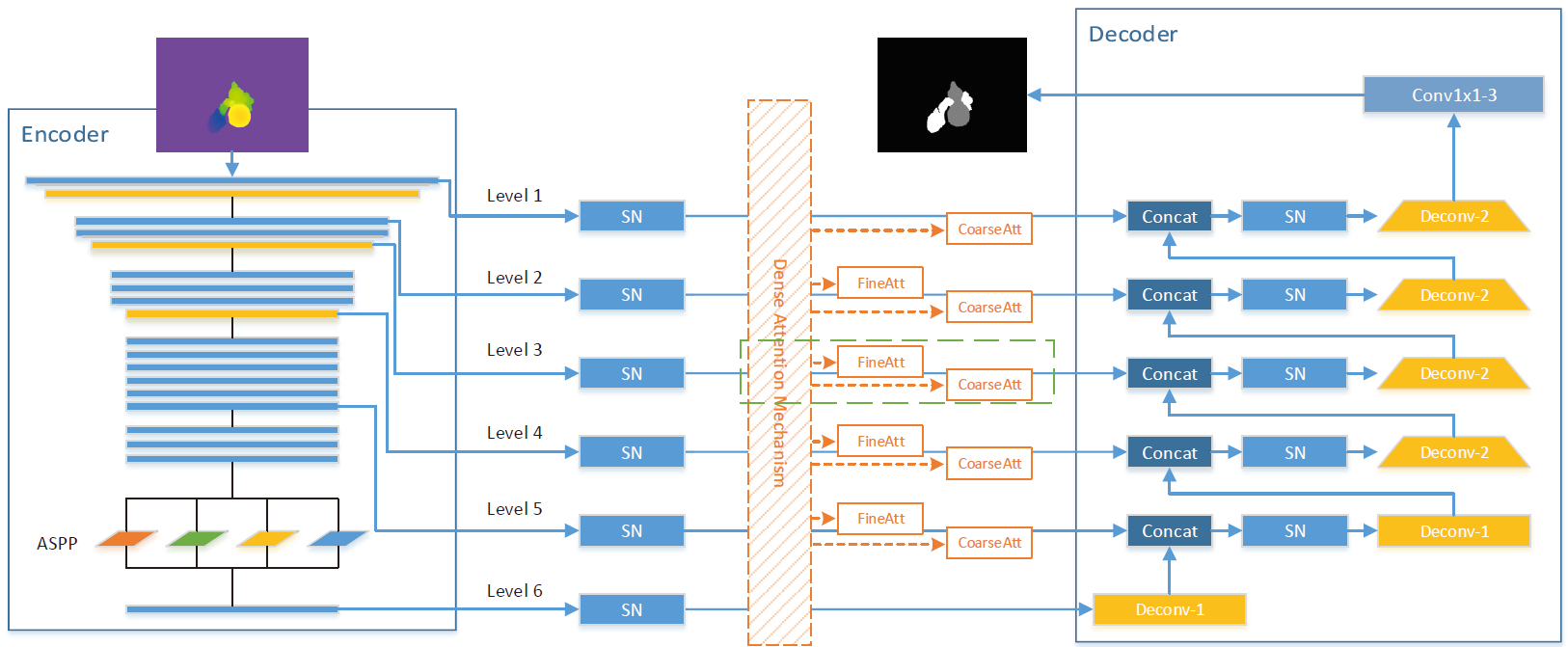}
  \end{center}
     \caption{DenseAttentionSeg model structure. The encoder is based on Deeplabv3 with Resnet-50 as backbone. The feature maps of the input depth image are extracted and split into different \emph{levels}. The Dense Attention Mechanism guides each level of feature maps with Fine and Coarse Attention, which is applied to the skip-connections in this figure. A decoder is applied to reconstruct the segmentation labels step by step. A detailed structure of the green dashed box is shown in Fig \ref{fig:attention}. \textbf{SN:} SqueezeNet. \textbf{Deconv:} deconvolution with stride 1 or 2. \textbf{Conv1x1-3:} final 1x1 convolution layer with softmax activation.}
  \label{fig:model}
\end{figure*}

As shown in Figure \ref{fig:model}, we propose DenseAttentionSeg to achieve real-time hand-object segmentation from depth input, which is based on a popular structure of encoder-decoder used in many segmentation tasks. To be specific, we use the same encoder structure in DeepLabv3+ \cite{chen2018encoder}. We also find it works well in our experiments. Meanwhile, we add a decoder that is powered by a densely connected attention structure to fuse all scales of features learned by the encoder and reconstruct the segmentation masks step by step.

\subsection{Encoder}
The encoder is based on DeepLabv3+ with stride-16 Resnet-50\cite{he2016deep} as the network backbone, together with Atrous Spatial Pyramid Pooling(ASPP). We do not use more powerful Resnet-101 and Xception as expressed in \cite{chen2018encoder} for the consideration of real-time performance. 

As the network goes deeper, the stride of feature maps gets bigger while the size gets smaller. We choose some intermediate feature maps as different \emph{levels}, which starts from \emph{level} 1 to \emph{level} $n$. Specifically, we select 5 intermediate feature maps in DeepLabv3, together with the final output of ASPP, thus $n=6$. 

\subsection{Dense Attention Mechanism}
\begin{figure}[h]
  \begin{center}
  \includegraphics[width=0.55\linewidth]{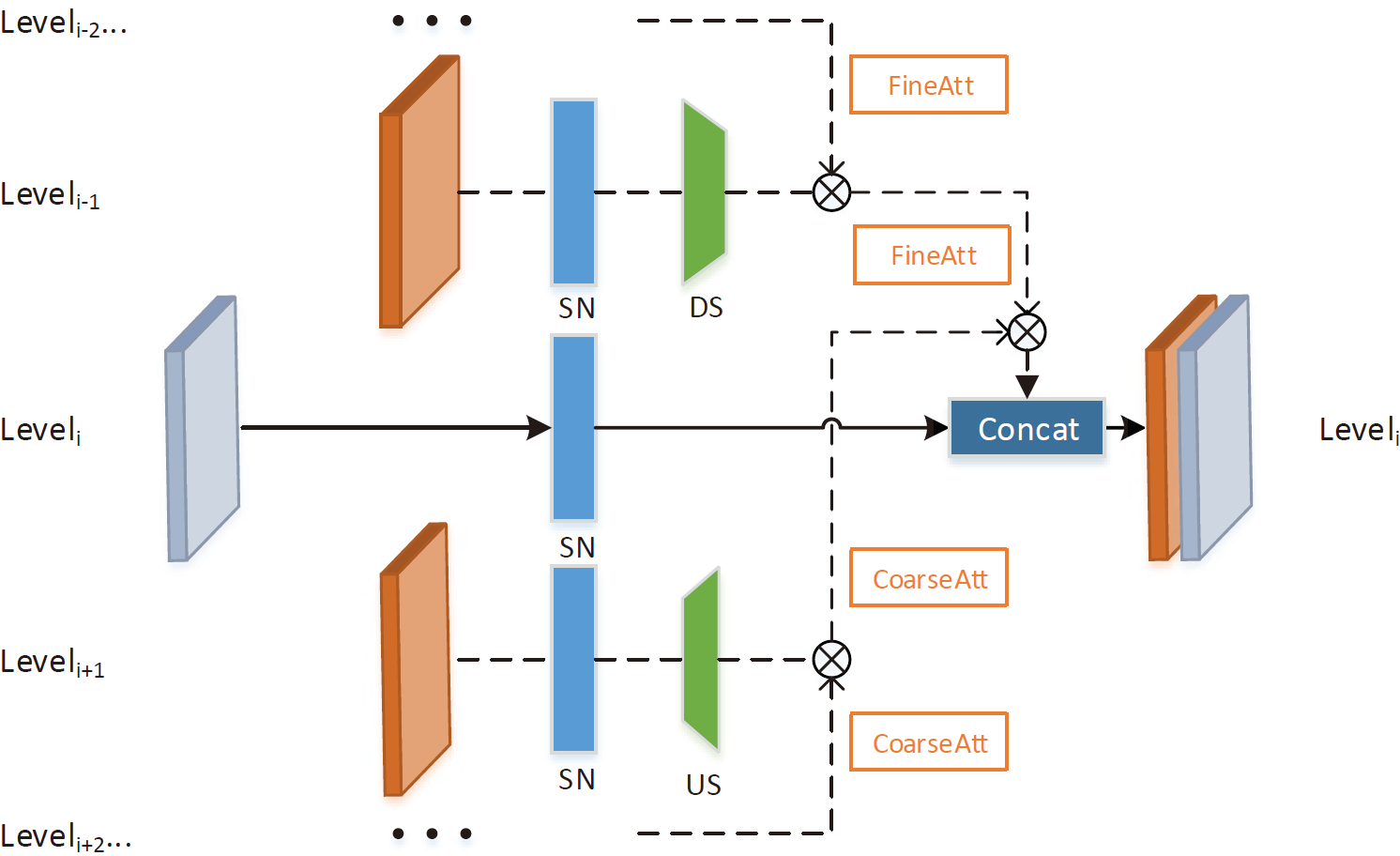}
  \end{center}
     \caption{Dense Attention Mechanism of \emph{level} i. Different \emph{levels} of attention map are gathered via element-wise multiplication. \textbf{SN:} SqueezeNet. \textbf{DS, US:} Bilinear down-sampling and up-sampling.}
  \label{fig:attention}
\end{figure}

We propose an attention mechanism to guide each \emph{level} to adjust features according to the information of different scales. Figure \ref{fig:attention} shows this mechanism in detail. We call it dense because each attention of every \emph{level} is fused from all the other \emph{levels}.

Assuming that we are manipulating \emph{level} $i$, the lower \emph{levels} of $i-1, i-2..., 1$ are gathered through element-wise multiplication step by step, and used to build the \emph{fine attention}, which conveys the shape and edge feature information to \emph{level} $i$. Since lower \emph{levels} have different number of feature channels and larger size than \emph{level} $i$, all the lower \emph{levels} are squeezed by a SqueezeNet and down-sampled by a bilinear down-sampling layer before the element-wise multiplication. Similarly, bilinear up-sampling is applied to \emph{level} $i+1, i+2..., n$ and then the \emph{coarse attention} is constructed, which is considered to convey semantic feature information.

The SqueezeNet can squeeze the channels of the input feature map to $k$, which not only reduces computation costs but also unifies channels for different \emph{levels}. A SqueezeNet contains a 1x1 and a 3x3 convolution layer, both with output channels of $k$.

\subsection{Decoder}
The decoder is applied to reform final segmentation label step by step. We adopt as many as 6 deconvolution\cite{zeiler2010deconvolutional} layers because we aim to segment hands in a close, fine scale and our operating image size is as big as $640\times480$. This structure is proved to have higher accuracy than those with fewer decoder layers on our dataset, namely Deeplabv3+.

\subsection{Loss Function}
We adopt \textbf{softmax cross entropy loss} as our basic training loss, as most of the other segmentation works do. 

\begin{equation}
    \label{softmax-loss}
    \mathcal{L}_{softmax}=-\sum_i y_i\log \frac{e^{x_i}}{\sum_j e^{x_j}} 
\end{equation}

In addition, we adopt a \textbf{contour loss} to refine the edges of hands and objects. First, extract edges of both logits of the network output and ground truth labels using Sobel operator, which can be accomplished via convolution layers. A Sobel operation is defined as follows:

\begin{equation}
    \label{equ:sobel}
    \mathcal{S}(M)=\sum_{l=1}^{n-1}\sqrt{(K_h*M_l)^2+(K_v*M_l)^2}
\end{equation}

Where $*$ indicates convolution operation, $M$ is the input map with $n$ channels representing different classes, and $M_l$ is the $l$th channel of $M$ indicating class $l$. We ignore background class whose label is 0 because it generates inconspicuous contours. The $K_h$ and $K_v$ are the horizontal and vertical kernel of the Sobel operator respectively. Then, blur the two contour maps using a Gaussian kernel, which can expand the influence of hand and object's edges smoothly as follows:

\begin{equation}
    \label{equ:gaussion-blur}
    \mathcal{B}(M)=G*M
\end{equation}

Where $G$ is a 5x5 Gaussian kernel with $\sigma=2.121$. The final contour loss is defined as an L2 loss below: 

\begin{equation}
    \label{equ:contour-loss}
    \mathcal{L}_{contour}=\frac{1}{N}\sum{(\mathcal{B}(\mathcal{S}(M_{labels}))-\mathcal{B}(\mathcal{S}(M_{logits})))^2}
\end{equation}

Where $M_{labels}$ is the ground truth labels and $M_{logits}$ is the output of our network. The equation sums up all pixels in all label classes and $N$ is the normalization parameter. The combined loss is computed as follows:

\begin{equation}
    \label{equ:joint-loss}
    \mathcal{L}_{finetuning}=\alpha \mathcal{L}_{softmax}+\beta\mathcal{L}_{contour}
\end{equation}

Where $\alpha$ and $\beta$ are the loss weights. We use it as a fine-tuning loss after the training of $\mathcal{L}_{softmax}$ finishes because the contour is not obvious at first. 

\subsection{Training Details}
During training, we apply data augmentation by random flipping and rotation. We also add a Gaussian noise to the input depth maps. The network is trained with a batch size of 4. We use Adam optimizer\cite{kingma2014adam} with an initial learning rate of 0.001, which decays by 0.8 every 80k steps to limit the highest value. The training is almost done after 100 epoch but we stop at 140 epoch to achieve every tiny improvement. Another 10 epoch is added to the fine-tuning procedure of the contour loss.

We set a high softmax loss weight of 5 for hand and object while keep that for the background as 1. The batch-norm loss weight is 0.001, and weights $\alpha$, $\beta$ in contour loss are 1.0 and 0.005. The output channel $k$ in SqueezeNet is set to 64.

\section{Experiments}
In this section, we first introduce the evaluation metrics. Then we quantitatively and qualitatively compare our method with the state-of-the-art models on the task of general semantic segmentation. Following, we evaluate the contour loss and the dense attention mechanism. After demonstrating the generalization and application of our technique, we introduce the current limitations of our model.

\subsection{Evaluation Metrics}
Our method aims to segment all pixels in a depth image into three classes - hand, object and background. To evaluate the method, we use the four metrics in \cite{long2015fully}, which are pixel accuracy, mean accuracy, mean IU and frequency weighted IU. Considering that our objective is to segment hands from objects, we also compute the mean IOU (Intersection over Union) for the hand and object classes.

\subsection{Segmentation Results}
We train different models on the training set of our InterSegHands and evaluate them on the validation set. The result is shown in Table \ref{table:qualitative-results}. Our DenseAttentionSeg with contour loss finetuning gains 93.94\% mean IOU, which is 1.18\%  higher than the baseline method DeepLabv3+(Resnet-50, OS-16)\cite{chen2018encoder}, and also outperforms other state-of-the-art methods, while still keeps high inference speed. 
\begin{table*}[ht]
  \begin{center}
  \begin{tabular}{c|c|cccc|c}
  \toprule
   & mean IOU & pixel acc. & mean acc. & mean IU & f.w. IU & time \\
  \midrule
  FCN-32s\cite{long2015fully} & 87.60 & 99.28 & 94.12 & 91.69 & 98.6 & 35.4ms \\
  FCN-16s\cite{long2015fully} & 87.59 & 99.39 & 94.48 & 91.67 & 98.83 & 35.6ms \\
  FCN-8s\cite{long2015fully} & 91.58 & 99.62 & 96.32 & 94.38 & 99.26 & \textbf{33.5ms} \\
  \midrule
  Large Kernel Matters (k=9)\cite{peng2017large} & 92.55 & 99.69 & 96.95 & 95.10 & 99.40 & 49.5ms \\
  RefineNet\cite{lin2017refinenet} & 92.53 & 99.64 & 96.74 & 94.98 & 99.30 & 61.9ms \\
  DeepLabv3+ (Resnet-50, OS-16)\cite{chen2018encoder} & 92.76 & 99.64 & 96.99 & 95.16 & 99.31 & 47.8ms \\
  \midrule
  DenseAttentionSeg & 93.65 & 99.74 & 97.48 & 95.77 & 99.50 & 53.7ms \\
  DenseAttentionSeg w/ CL & \textbf{93.94} & \textbf{99.75} & \textbf{97.67} & \textbf{95.99} & \textbf{99.51} & 53.7ms \\
  \bottomrule
  \end{tabular}
  \end{center}
  \caption{Quantitative results on InterSegHands validation set. The \textbf{mean IOU} is per class mean intersection over union rate for hand and object classes. The \textbf{time} is the inference time of batch size 1 on GTX Titan Xp with an input size of $640\times480$. \textbf{CL}: contour loss finetuning.}
  \label{table:qualitative-results}
\end{table*}

\begin{figure*}[htb]
  \begin{center}
  \includegraphics[width=0.95\linewidth]{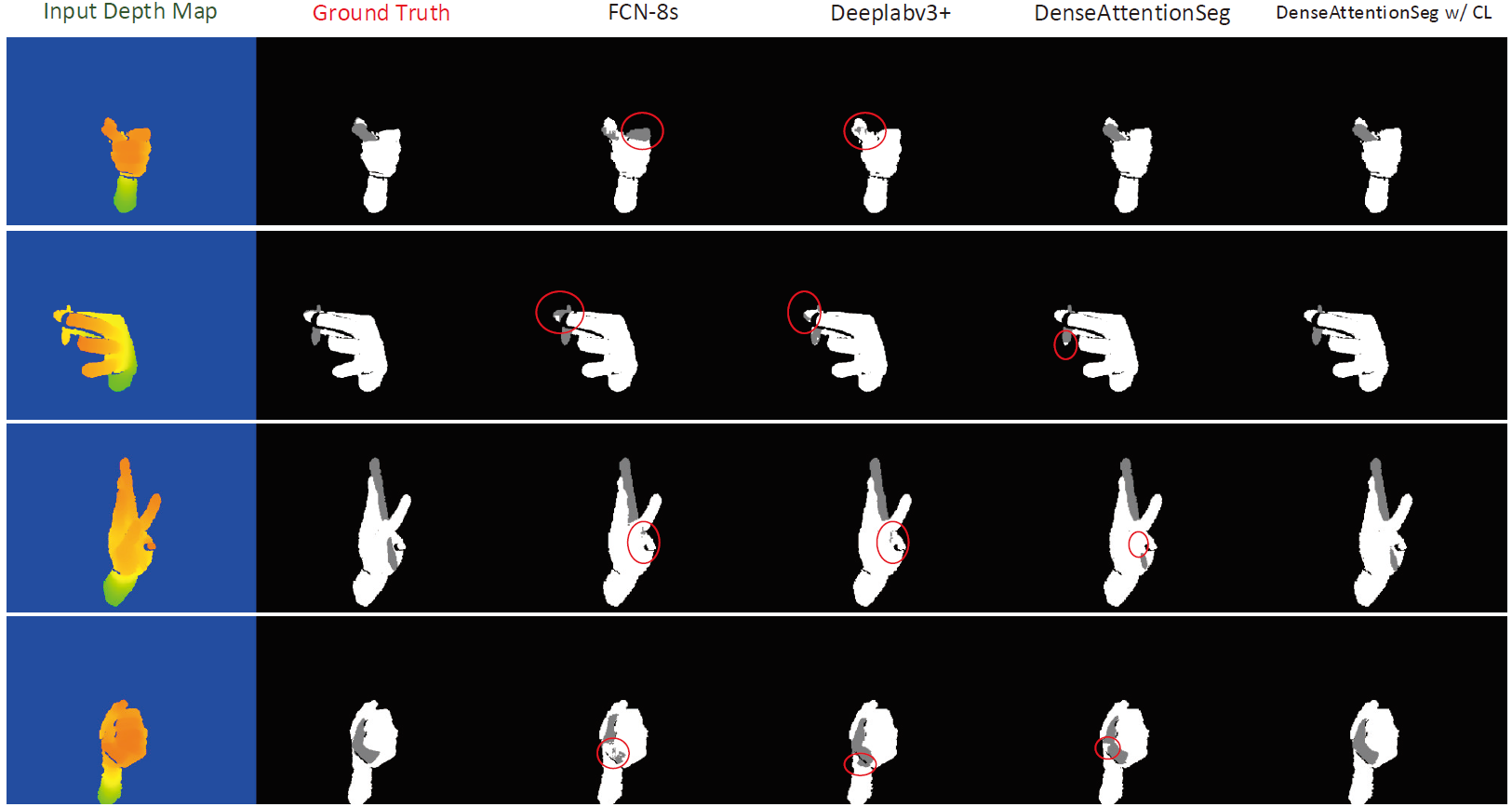}
  \end{center}
     \caption{Qualitative results on InterSegHands validation set. Our DenseAttentionSeg works better in more confusing and detailed regions, especially after contour loss finetuning.}
     \label{fig:results} 
\end{figure*}

Besides quantitative evaluations, we also perform some qualitative comparisons in Figure \ref{fig:results}. The light-weight FCN-8s fails to distinguish hand and object regions in some examples and cannot identify clear boundaries in some local regions. Deeplabv3+ is somewhat better but still generates noticeable errors. Our DenseAttentionSeg improves the result a lot. Even for small and heavily occluded objects, our model is still robust and generates promising results. 

\subsection{Evaluation on Dense Attention Mechanism}
\begin{figure*}[ht]
  \begin{center}
     \includegraphics[width=0.98\linewidth]{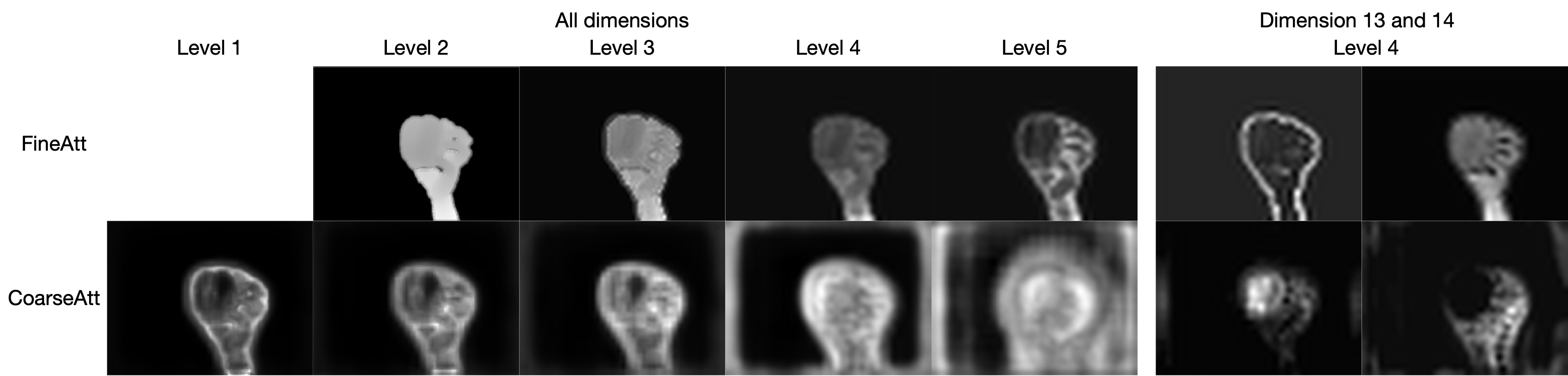}
  \end{center}
     \caption{Attention maps. On the whole, the fine attention is more detailed, while the coarse attention is more general. We can also see from the right that the coarse attention carries some semantic information, while the fine attention focuses on edge and shape as expected.}
  \label{fig:att}
\end{figure*}

Our dense attention mechanism aims to direct the skip-connections between encoder and decoder to merge high-\emph{level} semantic and low-\emph{level} detailed edge information. In our experiment, the dense mechanism provides a positive effect on our model. Specifically, when we drop the dense attention mechanism, our model gains a mean IOU of 92.81\%, which is still a little higher than our baseline model Deeplabv3+ due to our better feature reforming decoder. After adding the dense attention mechanism, the mean IOU rises by 0.84\%. The results are shown in Table \ref{table:attention-mechanism}. We also show the attention maps in Figure \ref{fig:att}.

\begin{table}[htbp]
  \begin{center}
  \begin{tabular}{ccc}
      \toprule
       & mean IOU & time \\
      \midrule
      Deeplabv3+ (baseline) & 92.76 & 47.8ms \\
      \midrule
      Ours w/ attention & 92.81 & 51.1ms \\
      Ours wo/ attention & 93.65 & 53.7ms \\
      \bottomrule
     \end{tabular} 
  \end{center}
  \caption{Evaluation on dense attention mechanism which increases the value of mean IOU from our baseline model.}
  \label{table:attention-mechanism}
\end{table}

\subsection{Evaluation on Contour loss}
\begin{figure}[t]
  \begin{center}
  \includegraphics[width=0.55\linewidth]{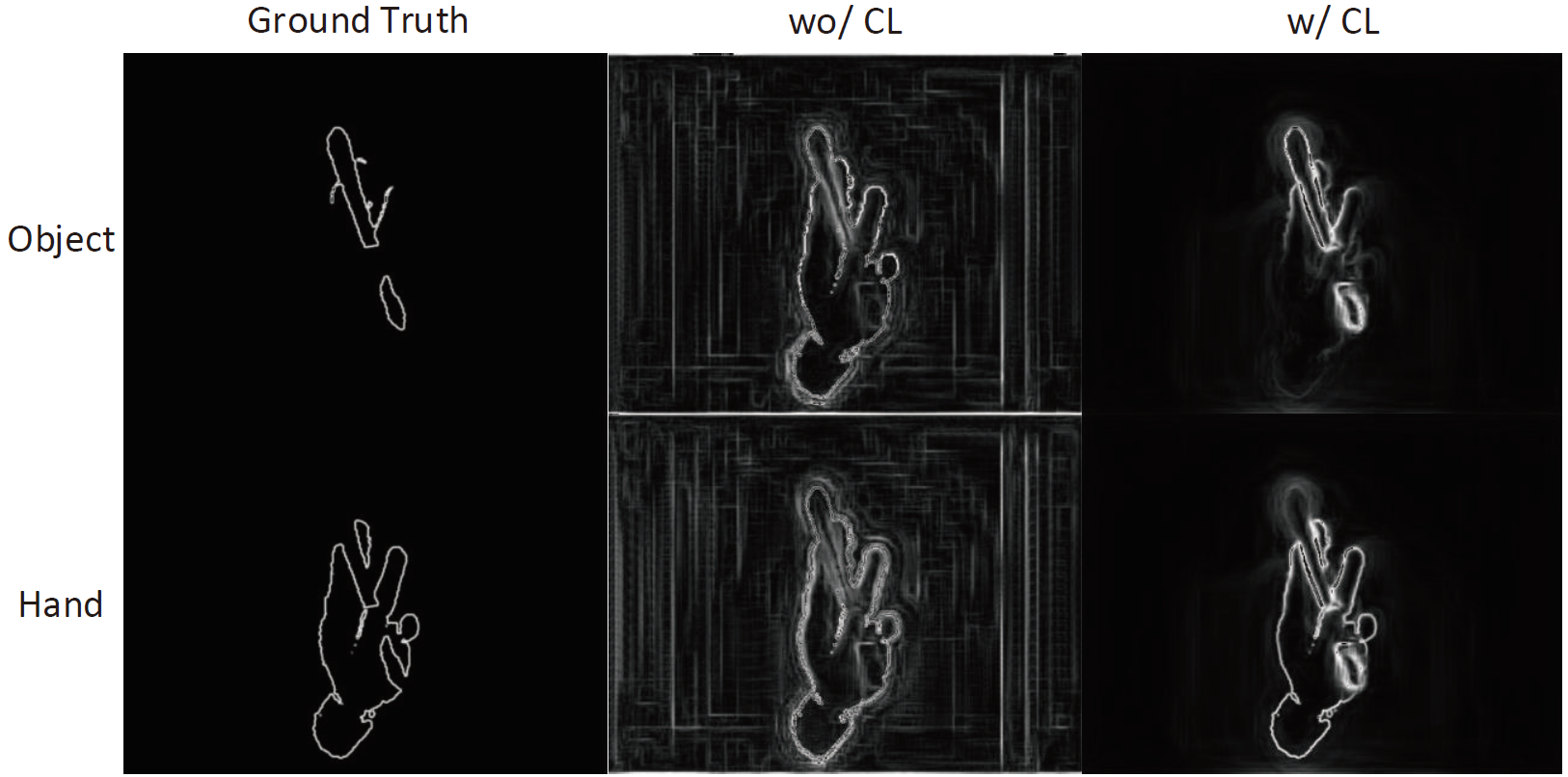}
  \end{center}
     \caption{Visualization of the contour loss extraction. With our contour loss finetuning, the model will get a smoother and better understanding of the boundaries of different classes.}
     \label{fig:contour} 
\end{figure}
The quantitative and qualitative comparison of our method with and without the contour loss are also shown in Table \ref{table:qualitative-results} and Figure \ref{fig:results}. We see that the contour loss further improves the result quality on all the metrics. Visually, we can see that the improvements are majorly lying on local boundaries, which is aligned with the theory. And those boundaries are important for accurate hand reconstruction and hand-based remote manipulation. We visualize the effect of our contour loss in Figure \ref{fig:contour}, where we perform boundary detection on the ground truth segmentation mask, the results without and with the contour loss. They are extracted as Eq. \ref{equ:contour-loss} does. 

\subsection{Generalization and Application}

\begin{figure}[ht]
  \begin{center}
  \includegraphics[width=0.55\linewidth]{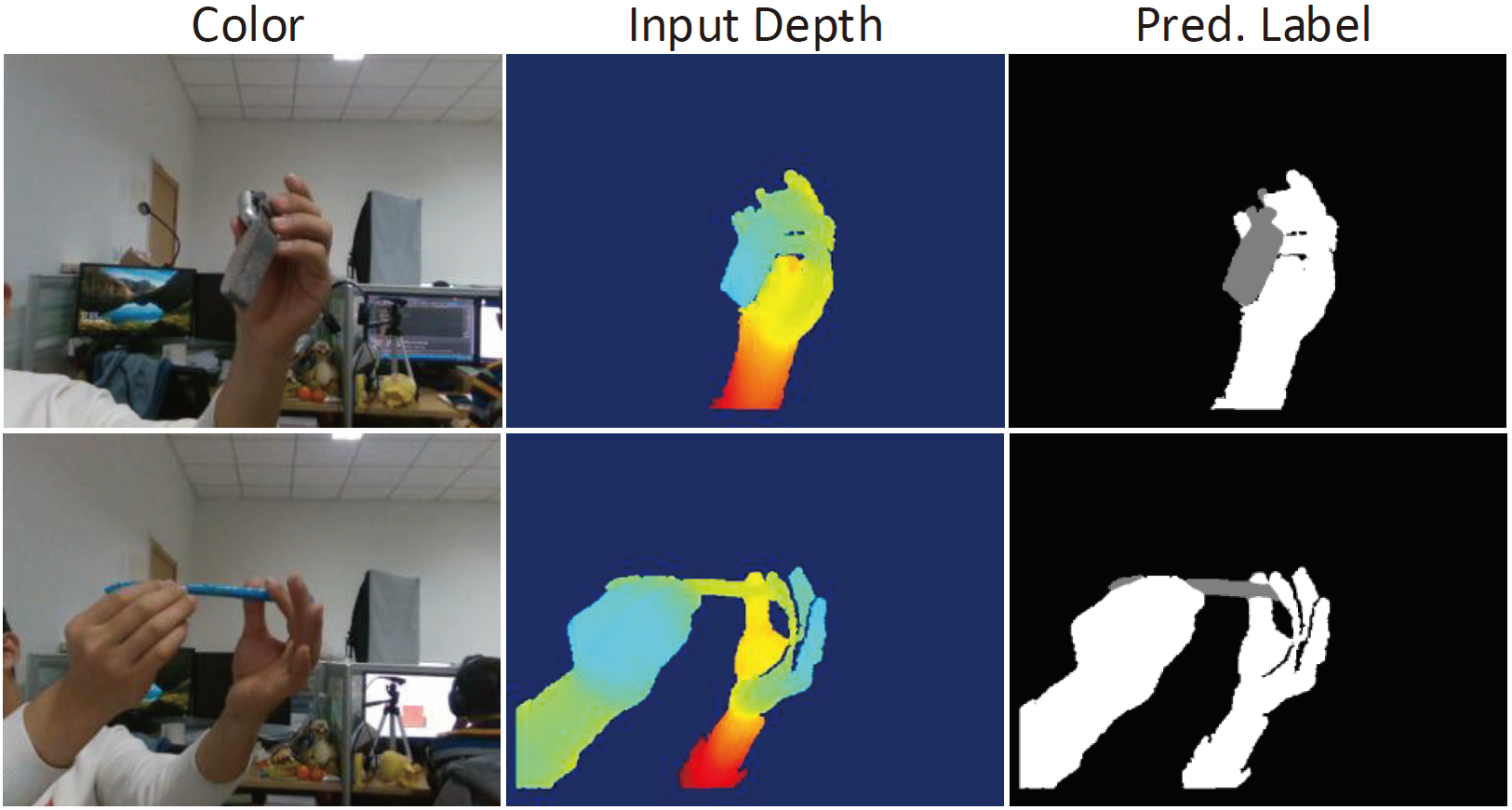}
  \end{center}
  \caption{Generalization. \textbf{First row:} non-rigid object that is out of training set. \textbf{Second row:} we can also handle both-hands cases.}
  \label{fig:generalization}
\end{figure}

We show some actual testing samples in Figure \ref{fig:generalization}. Our model structure and training strategy make it robust when applied to situations out of our dataset. 
\section{Conclusion}
We have proposed a 52K fine-scale hand-object segmentation dataset based on depth images - InterSegHands Dataset, and a deep model - DenseAttentionSeg to perform hand-object segmentation of interacting motions. Our DenseAttentionSeg contains a dense attention mechanism to fuse all \emph{levels} features as attention to guide the skip-connection between encoder and decoder, and a contour loss which helps to train the model to get more precise results on hand and object boundaries. According to our evaluations, our DenseAttentionSeg achieves the state-of-the-art performance among several advanced models on our dataset.

\bibliographystyle{unsrt}  
\bibliography{references}  






\end{document}